\documentclass{article}

\PassOptionsToPackage{square,numbers}{natbib}
\usepackage[preprint]{neurips_2020}
\usepackage[utf8]{inputenc}                             %
\usepackage[T1]{fontenc}                                %
\usepackage[colorlinks=true,citecolor=blue]{hyperref}   %
\usepackage{url}                                        %
\usepackage{booktabs}                                   %
\usepackage{amsfonts}                                   %
\usepackage{nicefrac}                                   %
\usepackage{microtype}                                  %
\usepackage{multirow}                                   %
\usepackage{amssymb}                                    %
\usepackage{graphics}                                   %
\usepackage{svg}                                        %
\usepackage{algorithm, algorithmic}%
\usepackage{afterpage}
\usepackage{lscape}
\usepackage{pdflscape}
\usepackage{courier} %
\usepackage{listings, xcolor}
\title{Universal Feature Selection Tool (UniFeat): An Open-Source Tool for Dimensionality Reduction}

\author{%
  Sina Tabakhi \\
  Department of Computer Science \\
  The University of Sheffield\\
  Sheffield, United Kingdom \\
  \texttt{stabakhi1@sheffield.ac.uk} \\
   \And
   Parham Moradi \\
   Department of Computer Engineering \\
   The University of Kurdistan \\
   Sanandaj, Iran \\
   \texttt{p.moradi@uok.ac.ir} \\
}

\begin{document}

\maketitle

\begin{abstract}
The Universal Feature Selection Tool (UniFeat) is an open-source tool developed entirely in Java for performing feature selection processes in various research areas. It provides a set of well-known and advanced feature selection methods within its significant auxiliary tools. This allows users to compare the performance of feature selection methods. Moreover, due to the open-source nature of UniFeat, researchers can use and modify it in their research, which facilitates the rapid development of new feature selection algorithms.
\end{abstract}

\section{Motivation and significance}
The field of data mining is concerned with knowledge discovery from data through the development of computer programs. The rapid advances in computer and database technologies have led to the production of high-dimensional data with large numbers of features in many fields. Most of the features are irrelevant and redundant, and these unnecessary features have stimulated a phenomenon in the data mining algorithms called `the curse of dimensionality' \cite{RN308,RN2,RN220}. Dimensionality reduction plays an essential role in addressing this issue by mitigating the dimensions of features while retaining the informative features of the original data. The dimensionality reduction techniques are generally performed based on either feature selection or feature extraction approaches \cite{khalid2014survey}. Feature selection is the process of identifying a subset of relevant features from the original feature set, while feature extraction methods transform the original data into a lower dimensional space to derive informative and non-redundant features. Feature selection is regarded as an important and active research area in data preprocessing, and numerous methods have been developed based on this technique \cite{solorio2020review}. Generally, feature selection methods are classified into three approaches: filter, wrapper, and embedded \cite{RN8,RN18,RN20}.

Several tools and libraries have been developed so far for the feature selection task. A collection of existing feature selection tools is presented and compared in Table \ref{table-1}. In this collection, we have incorporated open-source tools that have been recently updated, have been maintained for several years, or have been endorsed by the number of stars on their repository hosts (e.g., GitHub and SourceForge). Several vital metrics in the development of research software have been considered for comparing tools in Table \ref{table-1}:

\begin{itemize}
    \item Programming language: This metric reflects the programming languages most commonly used by developers in the feature selection research area.
    \item Documentation: This criterion demonstrates how well the software is documented for end-users and developers in terms of installation instructions, illustrative examples of the software's use cases, API documentation, and development tutorials to extend the software. 
    \item Graphical user interfaces (GUIs) presence: This feature shows how simple and quick the tool is to use for end-users.  
    \item Data format: This measure points out the tool’s support for a variety of input data formats.
    \item Creation and last update: These two criteria indicate the software's age and maintenance status.
    \item Main focus: This column specifies the primary goal of software development, which can be classified into two categories: data mining and feature selection tools. Data mining tools provide a general-purpose environment for machine learning models with different features such as data preprocessing, classification, regression, clustering, and visualization. In these tools, feature selection can only be considered a small module. Feature selection tools, on the other hand, are designed specifically for the feature selection task and offer a wide range of feature selection methods.
    \item Coverage of feature selection approaches: This indicator figures out the software's support for implementing baseline, well-known, and advanced feature selection methods in different categories. 
\end{itemize}

Table \ref{table-1} shows that the majority of existing software lacks a development guide, and several tools do not provide API documentation. Therefore, developers find it challenging to modify and extend the existing software. Another essential drawback with some tools, such as Weka \cite{RN76}, is that their documentation is not up-to-date to cover the most recent features and functionalities. Moreover, since non-expert researchers prefer to explore the software more effectively without coding requirements or command-line environments, providing a high-level representation of software functionalities via a GUI has become a standard in software development. Nonetheless, only a few of the available tools support a graphical interface for the users.

Another critical gap in Table \ref{table-1} that needs to be addressed is the tool's coverage of feature selection methods. Most existing tools are limited to a single feature selection approach, ignoring the others. Mlxtend \cite{raschkas_2018_mlxtend}, for example, is a data mining library whose feature selection module contains only baseline greedy wrapper-based methods. Another example is Weka, general-purpose data mining software that provides only a few conventional feature selection methods based on the filter and wrapper approaches. Besides, a small number of tools have incorporated all feature selection approaches in their codebase with the implementation of simple, baseline, and well-known methods. However, the community still demands new advanced feature selection techniques. RapidMiner \cite{RN229}, for instance, is an integrated platform for the generation of machine learning models in which many representative filter-based feature selection methods are implemented, but advanced ones are missing. Furthermore, only a few baseline methods based on the wrapper and embedded approaches are available in the RapidMiner repository. Only scikit-feature \cite{li2017feature} and ITMO\_FS \cite{pilnenskiy2020feature} have implemented numerous feature selection methods across all categories among the tools listed in Table \ref{table-1}. However, scikit-feature has not developed advanced feature selection methods in the wrapper and embedded classes, and its codebase has not been updated for several years. Moreover, ITMO\_FS has supplied several conventional wrapper feature selection methods, but the implementation of advanced techniques is still required.

Therefore, our aim in developing the Universal Feature Selection Tool (UniFeat) as a comprehensive feature selection tool includes seven aspects. (1) UniFeat implements well-known and advanced feature selection methods within a unified framework to respond to the community's pressing needs. (2) UniFeat can be regarded as a benchmark tool due to the development of methods in all the approaches. (3) The UniFeat functions provide essential auxiliary tools for performance evaluation, result visualization, and statistical analysis. (4) UniFeat has been completely implemented in Java and can be run on various platforms. (5) Researchers can utilize UniFeat through its GUI environment or as a library in their Java codes. (6) The open-source nature of UniFeat allows researchers to use and modify the tool to meet their research requirements and facilitate sharing their methodologies with the scientific community rapidly. (7) Finally, UniFeat includes a well-documented tutorial for developers and end users to support future software extensions.

\afterpage{%
\begin{landscape}
\begin{table}[!t]
\vspace*{-2cm}
\hspace*{1.5cm}
\small
  \centering
    \caption{Comparison of UniFeat to existing feature selection tools.}
    \renewcommand{\arraystretch}{1.7}
    \label{table-1}
    \makebox[\textwidth]{\resizebox{1.7\textwidth}{!}{\begin{tabular}{p{0.15\linewidth}p{0.12\linewidth}p{0.12\linewidth}p{0.2\linewidth}cp{0.2\linewidth}lllp{0.2\linewidth}p{0.2\linewidth}p{0.2\linewidth}}
    \toprule
    \multirow{ 2}{*}{\textbf{Tool name}} & \multirow{ 2}{*}{\parbox{2cm}{\centering \textbf{Programming language}}} & \multirow{ 2}{*}{\textbf{License}} & \multirow{ 2}{*}{\textbf{Documentation}} & \multirow{ 2}{*}{\textbf{GUI}} & \multirow{ 2}{*}{\textbf{Data format}} & \multirow{ 2}{*}{\textbf{Creation}} & \multirow{ 2}{*}{\textbf{Last update}} & \multirow{ 2}{*}{\textbf{Main focus}} & \multicolumn{3}{c}{\textbf{Feature selection approach}}  \\
    \cmidrule(r){10-12}
    & & & & & & & & & \textbf{Filter} & \textbf{Wrapper} & \textbf{Embedded} \\
    \midrule
    Weka \cite{RN76} & Java & GNU GPL & 
    Installation instructions \newline
    Illustrative examples \newline
    API documentation \newline
    Development guide & 
    \checkmark & ARFF, CSV, XRFF, XML & 1993 & 2022 & Data mining & 
    Few baseline methods are implemented &
    A small number of conventional methods are included &
    N/A \\
    
    RapidMiner \cite{RN229} & Java & GNU AGPL \newline A proprietary license  &
    Installation instructions \newline
    Illustrative examples \newline
    API documentation \newline
    Development guide  & 
    \checkmark & ACCDB, ARFF, CSV, DBF, DTA, HYPER, MDB, QVX, SAS, SAV, TDE, XLS/XLSX, XML, XRFF & 2001 & 2022 & Data mining & 
    Many representative methods are implemented, but advanced ones are missing &
    Few baseline methods are implemented &
    Few baseline methods are implemented\\

    Scikit-learn \cite{scikit-learn} & Python & BSD 3-Clause &
    Installation instructions \newline
    Illustrative examples \newline
    API documentation \newline
    Development guide & 
    N/A & CSV, XLS/XLSX, JSON, MAT, ARFF, SQL, Numpy arrays, LibSVM format & 2007 & 2022 & Data Mining & 
    A small set of simple baseline methods are provided & 
    Few sequential techniques are included & 
    Some baseline and well-known methods are involved \\
    
    Mlxtend \cite{raschkas_2018_mlxtend} & Python & BSD 3-Clause &
    Installation instructions \newline
    Illustrative examples \newline
    API documentation \newline
    Development guide & 
    N/A & CSV & 2014 & 2022 & Data mining & 
    N/A &
    A few greedy methods are implemented &
    N/A \\
    
    ITMO{\_}FS \cite{pilnenskiy2020feature} & Python & BSD 3-Clause & 
    Installation instructions \newline
    Illustrative examples \newline
    API documentation & 
    N/A & Scikit-learn input formats & 2018 & 2022 & Feature selection & 
    Many well-known and advanced methods are included &
    Several conventional methods are provided &
    A few advanced techniques are implemented \\

    FeatureSelector \cite{koehrsen2022selector} & Python & GNU GPL & 
    Illustrative examples \newline
    API documentation  & 
    N/A & CSV & 2018 & 2022 & Feature selection & 
    A few traditional methods are implemented &
    N/A &
    A few simple baseline techniques are involved \\
    
    Feature-engine \cite{Galli2021} & Python & BSD 3-Clause & 
    Installation instructions \newline
    Illustrative examples \newline
    API documentation \newline
    Development guide & 
    N/A & Scikit-learn input formats & 2019 & 2022 & Feature engineering and selection &
    Several conventional statistical methods are incorporated &
    A few traditional techniques are provided &
    A small number of baseline methods are involved \\
    
    Jx-WFST \cite{Too2022wfst} & MATLAB & BSD 3-Clause & 
    Illustrative examples & 
    N/A & MAT & 2020 & 2021 & Feature selection & 
    N/A &
    Many representative and advanced algorithms are implemented &
    N/A \\
    
    Mulan \cite{RN226} & Java & GNU GPL & 
    Installation instructions \newline
    Illustrative examples \newline
    API documentation \newline
    Development guide & 
    N/A & XML, ARFF & 2007 & 2020 & Multi-label learning & Few baseline methods have been implemented & N/A & N/A \\
    
    Feature Selection for Machine Learning \cite{Dutt2022featureselection} & Python & N/A & 
    Illustrative examples & N/A & Pandas input formats & 2018 & 2020 & Feature selection & 
    Some conventional statistical methods are provided &
    A small number of greedy algorithms are included &
    N/A \\
    
    FEAST \cite{RN232} & MATLAB\newline C/C++ & BSD 3-Clause & 
    Installation instructions \newline
    Illustrative examples & 
    N/A & MAT & 2011 & 2019 & Feature selection & Many standard mutual information-based methods are implemented & N/A & N/A \\
    
    Scikit-feature \cite{li2017feature} & Python & GNU GPL & 
    Installation instructions \newline
    Illustrative examples \newline
    API documentation & 
    \checkmark & MAT, CSV & 2015 & 2019 & Feature selection & 
    A range of well-known and advanced methods are included &
    A few greedy techniques are implemented &
    A few representative methods are provided \\
    
    FeatureSelect \cite{masoudi2019featureselect} & MATLAB & MIT & 
    Installation instructions \newline
    Illustrative examples  & 
    \checkmark & MAT, XLS, TXT & 2018 & 2019 & Feature selection & 
    A small number of baseline methods are included &
    A set of well-known and advanced methods are provided &
    N/A \\
    
    MLFeatureSelection \cite{Du2022mlfeatureselection} & Python & MIT & 
    Installation instructions \newline
    Illustrative examples  & 
    N/A & Pandas input formats & 2018 & 2019 & Feature selection & N/A & A small number of methods are implemented & N/A \\
 
    LOFS \cite{RN231} & MATLAB \newline OCTAVE & GNU GPL & 
    Installation instructions \newline
    Illustrative examples \newline
    API documentation  & 
    N/A & MAT, CSV & 2016 & 2016 & Online feature selection & N/A & Some online methods are involved & N/A \\
    
    \textbf{UniFeat} & Java & MIT & 
    Installation instructions \newline
    Illustrative examples \newline
    API documentation \newline
    Development guide  & 
    \checkmark & CSV & 2022 & 2022 & Feature selection & 
    Many well-known and advanced algorithms are implemented &  
    Some representative and advanced methods are included & 
    Several baseline and popular methods are provided \\
    
    \bottomrule
  \end{tabular}}}
\end{table}
\end{landscape}
\clearpage   %
}

\section{Software description}
\subsection{Software architecture}
UniFeat is an open-source Java tool for feature selection, principally developed at the University of Kurdistan, Iran. The goal of the project is to create a unified framework for researchers applying feature selection. For simplification of the development of the tool, UniFeat was divided into six main packages, including (1) featureSelection, (2) dataset, (3) classifier, (4) gui, (5) result, and (6) util, used for the following purposes.

\begin{enumerate}
  \item \textbf{featureSelection package}: provides all the feature selection methods that are implemented in the tool. This package is divided into three sub-packages to cover all the feature selection approaches. The current feature selection methods in the UniFeat repository are based on the filter, wrapper, and embedded approaches. The unified interface of the package allows researchers to implement their feature selection methods and share them with other researchers in the feature selection community.
  \item \textbf{dataset package}: is used for loading, saving, editing, and exporting different types of dataset files. 
  \item \textbf{classifier package}: collects several well-known and frequently used classifiers from the Weka software package. 
  \item \textbf{gui package}: provides GUIs that display the entire graphical representation of the panels for interaction with the user. Moreover, the package reports the results visually. It should be noted that this package has been separated from the others.
  \item \textbf{result package}: reports the performance results of feature selection methods based on different criteria such as accuracy and execution time.
  \item \textbf{util package}: presents various utility methods for manipulating arrays and matrices and performing basic statistical operations that can be used in the feature selection methods. 
\end{enumerate}

UniFeat is completely implemented in Java, and it can thus be run on any platform where Java Runtime Environment (JRE) is installed.

\subsection{Software functionalities}
\label{sec:func}
The open-source nature and structure of UniFeat bring about a set of opportunities that can be used by researchers to improve previous works and to enrich the repository of feature selection methods in UniFeat. 

The easiest way to use UniFeat is through its GUI. Some researchers use feature selection as a part of their own methods and prefer to embed the feature selection methods in their Java codes. Therefore, the design of UniFeat allows researchers to use UniFeat as a library in their own Java codes (additional details can be found in the UniFeat user manual at \url{https://unifeat.github.io/documentation.html}).

Besides, UniFeat supports two types of dataset files. In the first type, the dataset file consists of all the samples. In the second type, the datasets are initially divided into two portions: training and test sets. It should be noted that a specific way of representing datasets is needed for the tool; therefore, a simple preprocessing interface is provided to help users import datasets from different sources and convert them into the UniFeat format.

Once the feature selection process is completed in UniFeat, some necessary information is reported to the user in a plain text format. This includes information about the dataset, weights of features, classification accuracies, execution times, and subsets of selected features in different iterations. Furthermore, reduced datasets based on the subsets of selected features in various iterations are automatically created in two plain text formats: the comma-delimited format (i.e., the CSV file format) and the attribute-relation file format (i.e., ARFF, which is the standard format in Weka). These reduced datasets can easily be used for a fair comparison between the feature selection methods available in UniFeat and those implemented in any other software package.

The results obtained by UniFeat are visualized in the form of graphs, which can help users make better interpretations. After the feature selection process is complete, the user can see three graphs displaying execution time, accuracy, and error rate. Moreover, the values of the results obtained in each iteration and the average values in all the iterations can be reported in these graphs. To facilitate the reporting of the results, the user can save the graphs in the ``{\it png}'' image format.

To demonstrate that the experimental results are statistically significant, the Friedman test \cite{RN216} is currently provided in UniFeat for analysis of the results. The Friedman test is a non-parametric test used for the measurement of the statistical differences between methods over multiple datasets. 

\section{Illustrative examples}
This section provides three examples to describe the essential features of UniFeat. The first example demonstrates how to use UniFeat functionalities through its GUI, while the second illustrates how to use UniFeat APIs inside the Java code. The final case study explains how to carry out the Friedman test in UniFeat.

\begin{figure}[!ht]
  \centering
  \includegraphics[width=6cm]{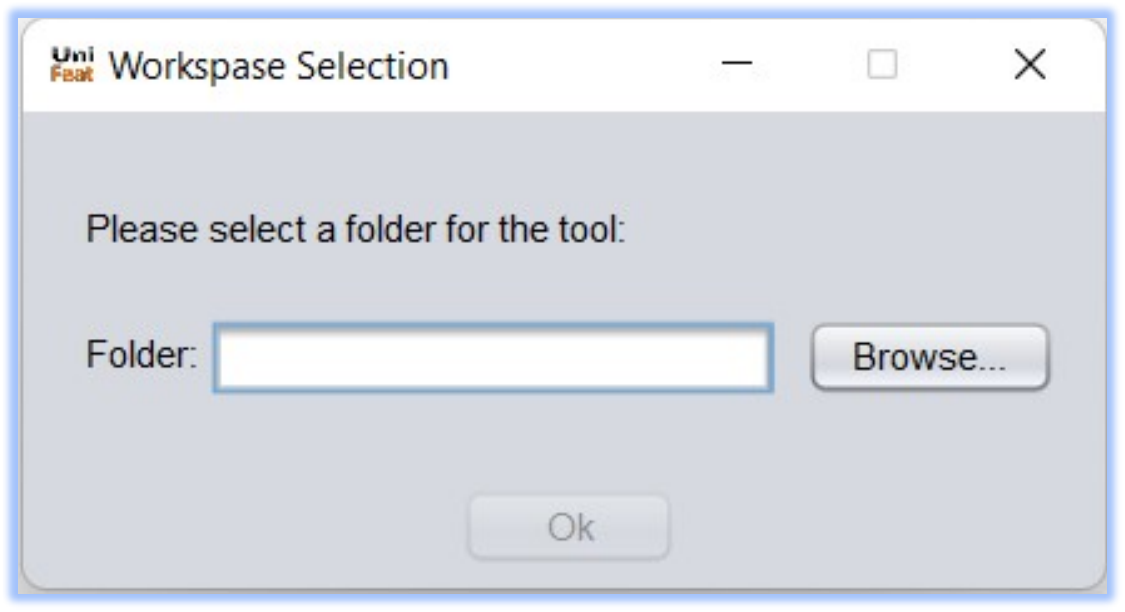}
  \caption{The workspace selection panel.}
  \label{figure_1}
\end{figure}

\begin{figure}[!ht]
  \centering
  \includegraphics[width=9.8cm]{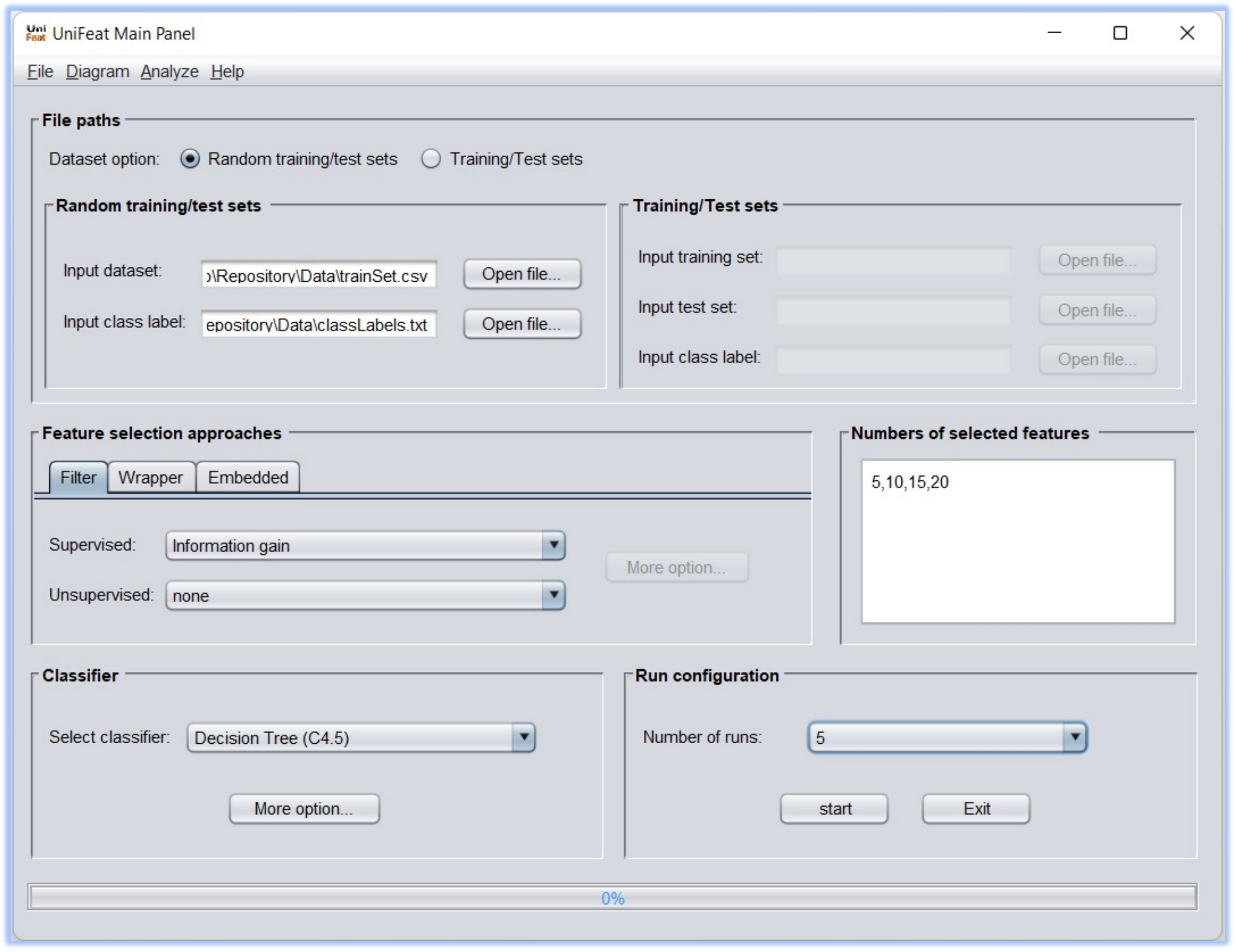}
  \caption{The main panel of UniFeat.}
  \label{figure_2}
\end{figure}

Figure \ref{figure_1} depicts a starting point for launching UniFeat's initial GUI window, which is used to select a workspace for the tool. After selecting a path, the user will see a new GUI window, which is the main panel of UniFeat (see Figure \ref{figure_2}). This panel provides access to all the tool facilities using form filling and menus. There are five different parts in the main panel corresponding to the specific tasks of the tool. (1) The ``{\it File paths}'' panel allows users to load UniFeat-formatted dataset files for future purposes. Datasets can be easily imported into the tool as CSV files. The dataset option is provided to support two alternative types of dataset files, as described in Section \ref{sec:func}. (2) In the ``{\it Feature selection approaches}'' panel, well-known and advanced feature selection techniques are involved, which can be chosen through three different tabs, including the ``{\it Filter},'' ``{\it Wrapper},'' and ``{\it Embedded}.'' Since many feature selection methods have adjustable parameters that need to be set, the tool includes a ``{\it More option…}'' button for configuring them. (3) The ``{\it Numbers of selected features}'' is designed to enable users to enter the various numbers of selected features altogether. Commas must separate these values. Figure \ref{figure_2} indicates how the given feature selection method should select 5, 10, 15, and 20 features. (4) In the ``{\it Classifier}'' panel, a classifier can be selected to evaluate the subsets of selected features. The Weka software package \cite{RN76} has been utilized for the implementation of the classifiers. In addition, the ``{\it More option…}'' button is embedded in this panel to adjust the parameters of the classifiers. (5) The ``{\it Run configuration}'' panel is incorporated to repeat the feature selection process multiple times in order to reduce the impact of randomness and produce stable results.

\begin{figure}[!ht]
  \centering
  \includegraphics[width=9.8cm]{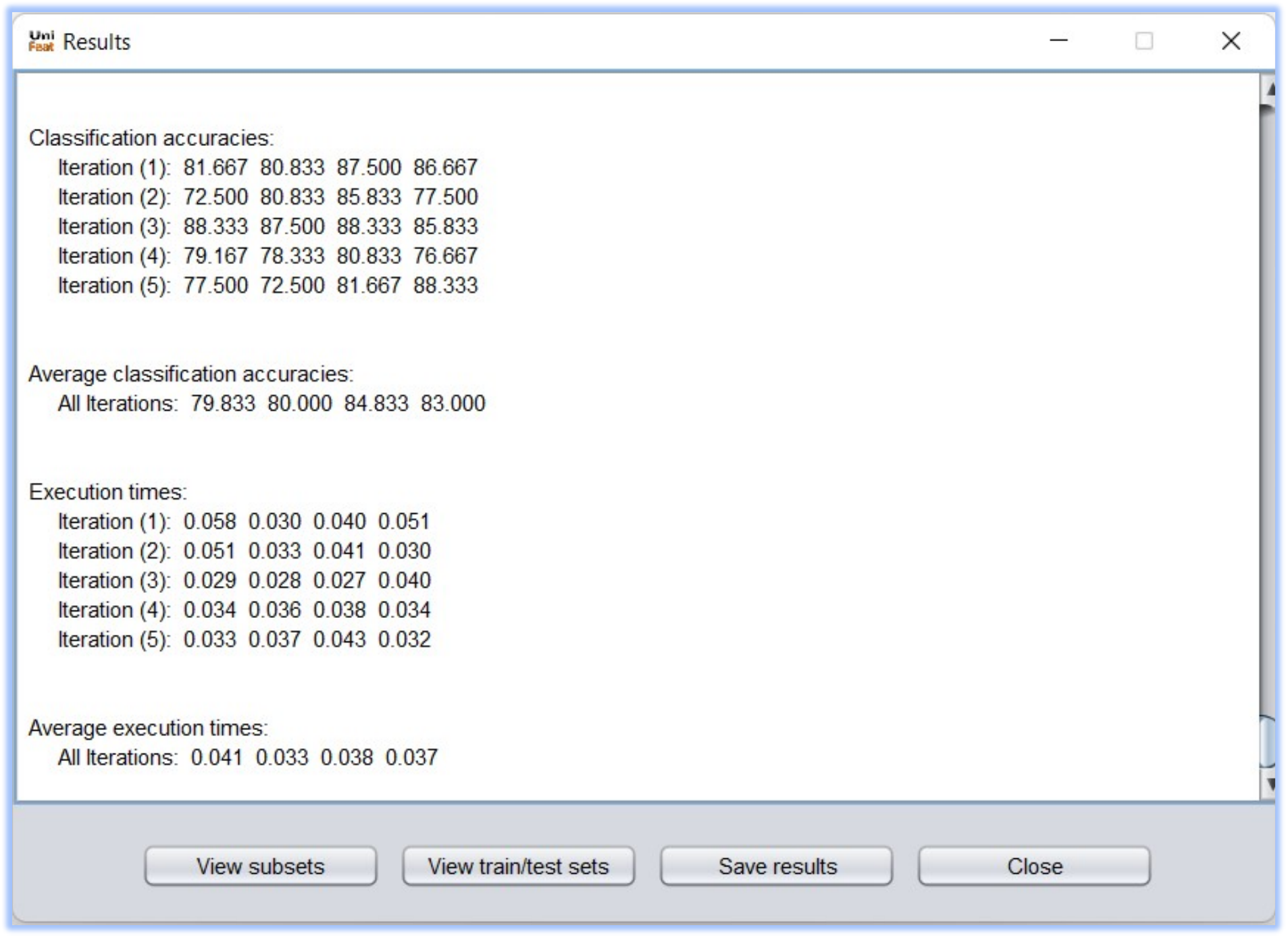}
  \caption{An example of the resulting interface.}
  \label{figure_3}
\end{figure}

\begin{figure}[!ht]
  \centering
  \includegraphics[width=13cm]{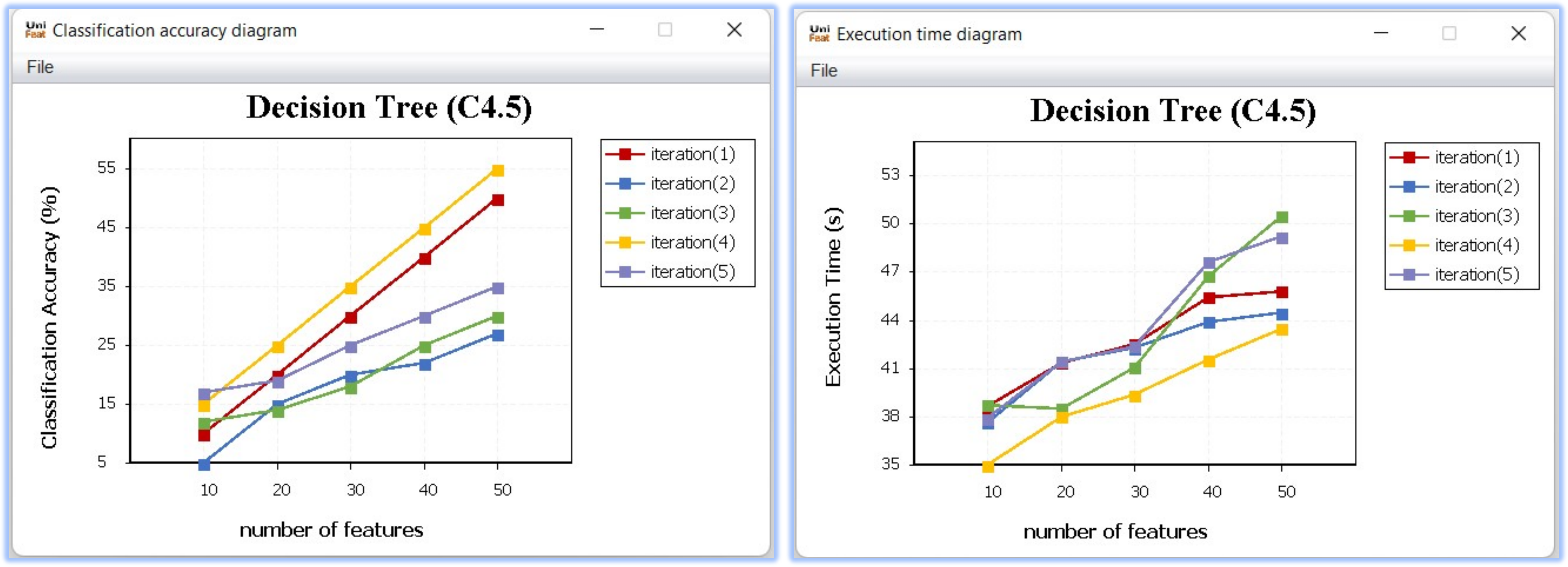}
  \caption{The diagrams of the results produced by UniFeat.}
  \label{figure_4}
\end{figure}

The user can initiate the feature selection process by clicking the ``{\it Start}'' button. If all the tool's requirements are met, the user will see the results in a variety of formats. Figure \ref{figure_3} shows an example of UniFeat output results. The resulting interface reports various useful information to the user, including information about the dataset, weights of features, classification, accuracy values, execution times, and subsets of selected features in each iteration. Moreover, visualizing the results in the form of diagrams is an additional representation of the results that can help users obtain more precise interpretations (see Figure \ref{figure_4}). UniFeat's user manual is available at \url{https://unifeat.github.io/documentation.html} for more information.

Algorithm \ref{algorithm-1} demonstrates a simple Java program that performs the feature selection using the information gain method \cite{RN66} implemented in UniFeat. In the first step (lines 9 and 10), the dataset files are loaded using invoking appropriate functions for future purposes. In this example, the datasets are initially divided into training and test sets, both of which must be provided in the UniFeat format. Then lines 12 through 18 show some helpful information about the dataset. Next, the feature selection process is performed by applying the information gain technique in lines 20-31. The subset of selected features is then displayed to users between lines 33 and 37. Finally, lines 45-51 provide how the reduced datasets are generated in the {\it .csv} and {\it .arff} formats considering the subset of selected features.

\begin{algorithm}[!ht]
  \centering
  \caption{Example code for feature selection by invoking UniFeat APIs.}
  \label{algorithm-1}
\begin{lstlisting}[language=Java]
   import unifeat.dataset.DatasetInfo;
   import unifeat.featureSelection.filter.supervised.InformationGain;
   import unifeat.util.FileFunc;

   public class Main {
    
       public static void main(String[] args) {
           //reading the datasets files
           DatasetInfo data = new DatasetInfo();
           data.preProcessing("data/trainSet.csv", "data/testSet.csv", "data/classLabels.txt");

           //printing some information of the dataset
           int sizeSelectedFeatureSubset = 2;
           System.out.println(" no. of all samples : " + data.getNumData()
                   + "\n no. of samples in training set :  " + data.getNumTrainSet()
                   + "\n no .of samples in test set : " + data.getNumTestSet()
                   + "\n no. of features : " + data.getNumFeature()
                   + "\n no. of classes : " + data.getNumClass());

           //performing the feature selection by information gain method
           InformationGain method = new InformationGain(sizeSelectedFeatureSubset);
           method.loadDataSet(data);

           String message = method.validate();
           //checking the validity of user input values
           if (!message.isEmpty()) {
               System.out.print("Error!\n  " + message);
           } else {
               method.evaluateFeatures();
               int[] subset = method.getSelectedFeatureSubset();
               double[] infoGainValues = method.getFeatureValues();

               //printing the subset of selected features
               System.out.print("\n subset of selected features: ");
               for (int i = 0; i < subset.length; i++) {
                   System.out.print((subset[i] + 1) + "  ");
               }

               //printing the information gain values
               System.out.println("\n\n information gain values: ");
               for (int i = 0; i < infoGainValues.length; i++) {
                   System.out.println(" " + (i + 1) + " : " + infoGainValues[i]);
               }

               //creating reduced datasets as the CSV file format
               FileFunc.createCSVFile(data.getTrainSet(), subset, "data/newTrainSet.csv", data.getNameFeatures(), data.getClassLabel());
               FileFunc.createCSVFile(data.getTestSet(), subset, "data/newTestSet.csv", data.getNameFeatures(), data.getClassLabel());

               //creating reduced datasets as the ARFF file format
               FileFunc.convertCSVtoARFF("data/newTrainSet.csv", "data/newTrainSet.arff", "data", sizeSelectedFeatureSubset, data);
               FileFunc.convertCSVtoARFF("data/newTestSet.csv", "data/newTestSet.arff", "data", sizeSelectedFeatureSubset, data);
           }
       }
   }
\end{lstlisting}
\end{algorithm}

\begin{figure}[!ht]
  \centering
  \includegraphics[width=9cm]{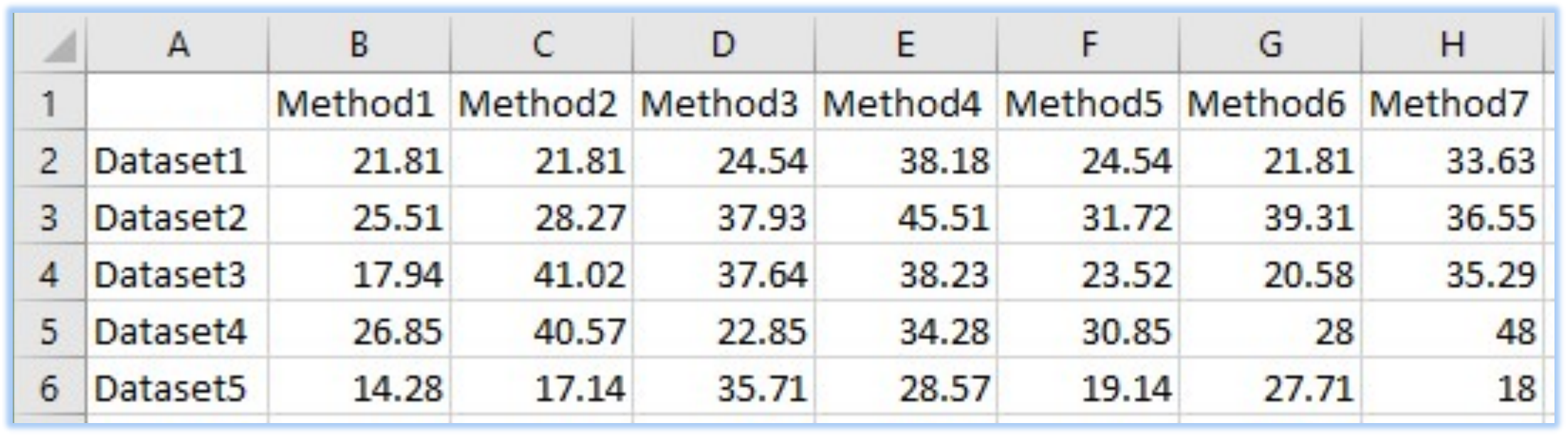}
  \caption{An example of the results prepared for conducting the Friedman test in UniFeat.}
  \label{figure_5}
\end{figure}

\begin{figure}[!ht]
  \centering
  \includegraphics[width=10.4cm]{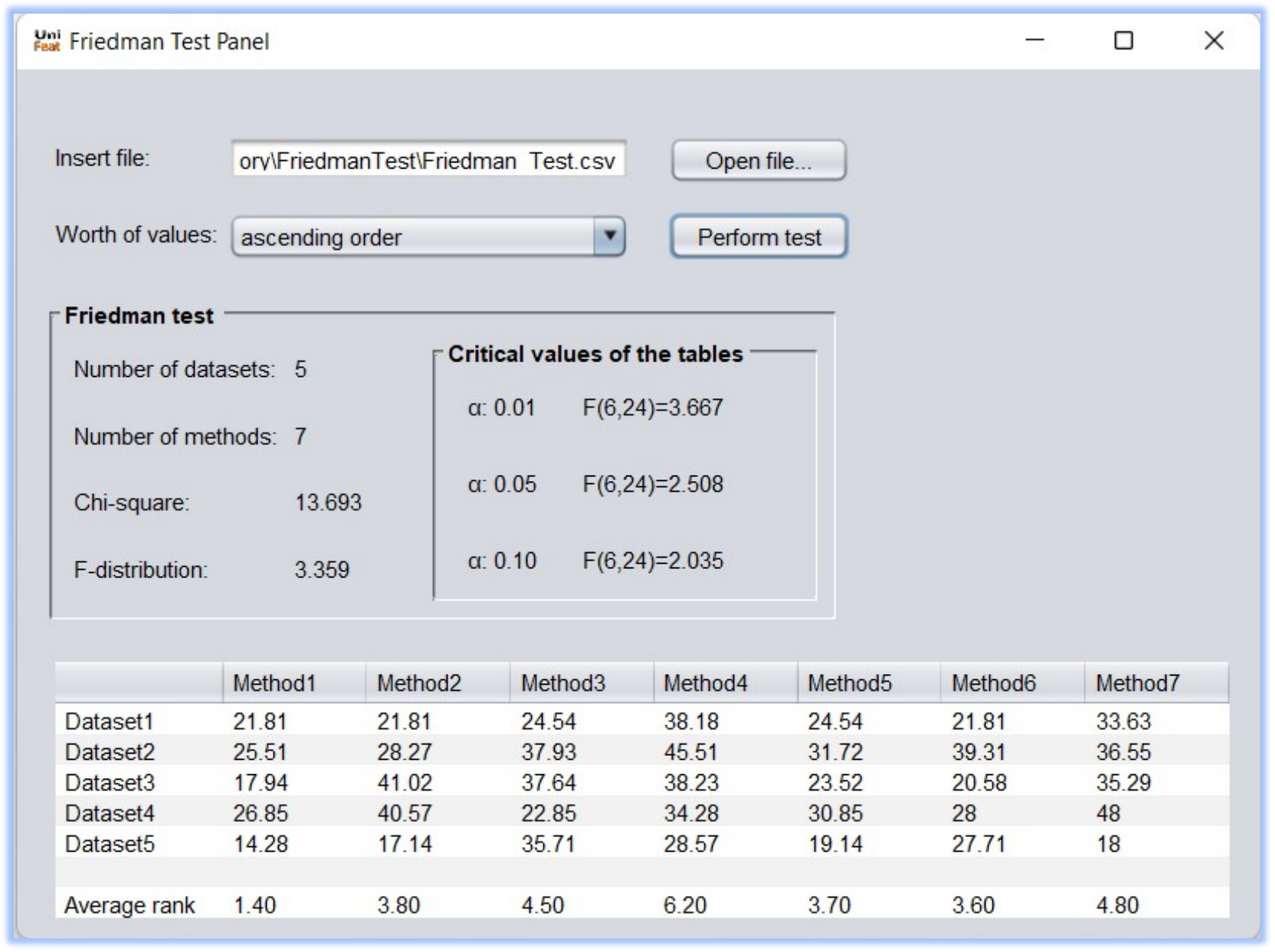}
  \caption{The Friedman test panel.}
  \label{figure_6}
\end{figure}

The final example explains how to apply the Friedman test in UniFeat to statistically analyze the performance of feature selection algorithms. Given that the performance of multiple feature selection methods has been evaluated in terms of classification error rate over numerous datasets. Figure \ref{figure_5} summarizes the obtained results from seven feature selection methods applied to the five datasets. The Friedman test in UniFeat can be carried out to prove that the results are statistically significant. After importing the file presented in Figure \ref{figure_5} into the ``{\it Friedman Test Panel}'' and executing the test, several helpful information will be reported to users, including the average values of each method over all datasets, Chi-square and F-distribution values, and critical values of the table based on various significant levels (i.e., $\alpha$ parameter). The results of the Friedman test conducted on the prepared file are illustrated in Figure \ref{figure_6}. It can be seen that when the $\alpha$ is equal to 0.05 and 0.1, the F-distribution value (i.e., 3.359) is greater than the critical values of the table. Consequently, it can be proved that the results are statistically significant, and method 1 has the best performance compared to other methods. Moreover, the value of F-distribution is less than the critical value when $\alpha$ = 0.01, indicating that the performance of feature selection methods is equivalent.

\section{Conclusions}
The UniFeat tool was developed with the purpose of implementing well-known and advanced feature selection methods within the auxiliary tools for researchers. It helps researchers in different fields needing a feature selection technique to reduce the dimensionality of data. The tool's graphical user interface can make it easy for users to perform the feature selection process. Moreover, the design of UniFeat as a {\it jar} file can be helpful for researchers who wish to employ feature selection in their own Java codes. UniFeat has also been documented for end-users to use the software easily and for developers to further modification and extension. UniFeat is an open-source tool distributed under the MIT License terms.

The developers of UniFeat will try to constantly add new advanced feature selection methods to the tool in order to enrich its codebase. Another extension would be to develop new statistical analyses in UniFeat. Moreover, the visual displays will be updated to support additional visual outputs.

\begin{ack}
We are grateful to Mr. Shahin Salavati, who implemented most of the GUIs. We also thank the professors who provided us with feedback about the tool, especially Dr. Fardin Akhlaghian at the University of Kurdistan and Dr. Haiping Lu at the University of Sheffield.
\end{ack}

\medskip

\small

\bibliography{neurips_2020}

\end{document}